\documentclass[10pt,twocolumn,letterpaper]{article}

\usepackage{iccv}
\usepackage{times}
\usepackage{epsfig}
\usepackage{graphicx}
\usepackage{amsmath}
\usepackage{amssymb}
\usepackage{booktabs}
\usepackage{float}

\usepackage[breaklinks=true,bookmarks=false]{hyperref}

\iccvfinalcopy 

\def\iccvPaperID{****} 

\ificcvfinal\fi

\begin{document}

\title{Learning to Localize Temporal Events in Large-scale Video Data}

\author{Mikel Bober-Irizar\\
University of Cambridge\\
{\tt\small mikel@mxbi.net}
\and
Miha Skalic\\
University Pompeu Fabra\\
\and
David Austin\\
Intel Corporation\\
{\tt\small david.j.austin@intel.com}
}

\maketitle
\ificcvfinal\thispagestyle{empty}\fi

\begin{abstract}

\noindent   We address temporal localization of events in large-scale video data, in the context of the Youtube-8M Segments dataset.  This emerging field within video recognition can enable applications to identify the precise time a specified event occurs in a video, which has broad implications for video search.  To address this we present two separate approaches: (1) a gradient boosted decision tree model on a crafted dataset and (2) a combination of deep learning models based on frame-level data, video-level data, and a localization model. The combinations of these two approaches achieved 5th place in the 3rd Youtube-8M video recognition challenge.
   
\end{abstract}

\section{Introduction}

Understanding the contents of a scene from its video and audio features is a central capability of AI based systems today.  There are many algorithms available to perform a classification task when presented with such input features.  What is not as common but perhaps as valuable if not more so is to be able to take an long video and search for a given labeled scene.  This scenario is much more common in the real world applications where video editing is costly and global data and meta-data can be noisy.  Practical use cases for such kind of applications could include but not limited to: (1) police forces searching for "use of force" videos out of a large video library to properly asses and train their employees, (2) a security firm identifying hazards or threats from a repository of video, or (3) a video hosting service allowing users or advertisers to specifically identify scenes of interest of their users from a large database. Some recent approaches to localizing temporal events in a video include detecting audio-video events with attention\cite{av_event_local} and using a modified R-CNN architecture to localize temporal context of actions\cite{fasterrcnn}.

Our contribution is to demonstrate that applying traditional machine learning and deep learning approaches to pre-extracted features is capable of achieving relatively high performance without the need for more expensive action recognition networks or deep ensembles of many diverse networks.

\section{Dataset \& Competition Setup}

The 3rd Youtube-8M video recognition challenge builds on an updated version of the original Youtube-8M dataset\cite{2016dataset}, known as the Youtube-8M segments dataset. This dataset contains 6.1 million YouTube videos, each labelled with some of 3862 classes (on average 3 per video). Due to dataset size constraints, instead of providing full audiovisual streams, each video is sampled at 1 frame per second (up to a maximum of 300 seconds), and each frame is passed though pre-trained Inception-v3\cite{inceptionv3} network as well as an acoustic model\cite{acousticmodel}, followed by PCA and quantization to generate a pre-extracted 1152-D feature vector for each frame. This leads to a dense 2 dimensional matrix of features for each video in the dataset. The first two Youtube-8M competitions\cite{2018challenge} were built upon this dataset, and so there is much existing research into solutions for the video-level classification problem.\cite{2017best}\cite{mikelyt}\cite{2018best}

In addition to the above data, Youtube-8M segments adds 237,000 human-labelled \textit{5 second segments} within videos - each labelled segment either contains or does not contain a specific class (out of a subset of 1000 classes). Fewer than 50,000 (1\%) of the videos in the dataset are labelled in this way, with each of these videos containing on average 5 labelled segments. This yields a segment-level labelling which is \textbf{extremely sparse}, with only 0.0001\% of all possible segment-class pairs being labelled, and fewer than 250 total segment labels per class.

The goal of the 3rd Youtube-8M video recognition challenge is to predict \textit{segment-level} class labels for each video in the test set. Given the scarcity of segment-level training examples, it is vital for any solution to leverage the vast number of video-level training examples in some way during training - the problem is one of weakly supervised learning. In essence, we have to learn to localise classes within videos without knowing (or rarely knowing) where the classes appear in the training set examples.

\section{Evaluation}
The Kaggle competition is evaluated using Mean Average Precision (MAP@K) where $K=100,000$. The Average Precision (eq. \ref{eq1}) is computed on a \textit{per class} basis, and averaged across the 1000 classes in the dataset. 

\begin{equation}
    \label{eq1}
    MAP@K = \frac{\sum^{K}_{k=1}{P(k) \times rel(k)}}{N_{pos}} ,
\end{equation}

\noindent where $P(k)$ is the precision in the first k predictions, $rel(k)$ is 1 if prediction $k$ is correct and 0 otherwise, and $N_{pos}$ is the total number of positive labelled segments for that class.

Thus for each class, the task is to predict the top 100,000 labelled video segments which may contain that class in order of likelihood. As the segment labelling is also extremely sparse in the test set (and we do not know which segments are labelled), it is necessary to predict many more than 100,000 segments for each class in order for 100,000 predictions to be `considered'. In practice, the limiting factor on number of predicted segments is the timeout on Kaggle's evaluation servers (which will fail if a submission takes more than a few minutes to score) rather than the limit of $K=100000$.

\section{Video-level Baseline}
\label{sec:baseline}

Because the evaluation metric is evaluated on a per-class basis, and not per-video (for example, ranking segments within videos), this means that it is possible to achieve a high score simply by calculating a likelihood for each (video, class) pair - as in previous competitions - and assuming that within each video, all segments have the same probability of containing that class. This yields a pseudo-ranking for the segments which can be submitted.
We took the winning model from the 2018 Youtube-8M competition\cite{2018best} (with a GAP score above 0.890), and used the naive approach detailed to achieve a score of \textbf{0.68182 MAP}.

Thus, it is important to note that a large part of the evaluation metric for this dataset still relies on ordering videos based on likelihood of each label, in addition to ranking segments within each video, and a good solution must be able to effectively do both.

To this end, here we present and contrast two machine learning approaches that have been developed to localize target events in videos. The first approach utilizes gradient boosting decision trees, while the second approach is a more `conventional' deep learning-based approach.  These two methods are complimentary, trained in parallel, and are subsequently combined to produce our final approach.

\section{Gradient Boosting Trees for Event Localization}
\label{sec:gbm}

\subsection{Frame-Level Model}

First, to leverage the large number of \textit{video-level} labels available (7 million training examples), a basic feed-forward neural network was trained to predict labels based on the provided feature vectors averaged across the time dimension to form a single 1152-D feature vector per video. This network was trained on video-level average data (with the available video-level labels), but is then used to predict on \textbf{individual frames} instead. We refer to a prediction of likelihoods from this model on frame $X \in \mathbb{R}^{1152}$ as $fl(X) \in \mathbb{R}^{1000}$.

\subsection{Frame-level dataset construction}

Next, for each of the 1000 classes: a dataset of frames was constructed by taking all the labelled 5-second segments for that class, and sampling 9 frames from each segment (two on either side of the labelled five, assuming they also have the same class), and concatenating these segments to form a frame-level dataset for each class. This approach yielded approximately 1000 ``definitely positive" and 1000 ``definitely negative" frames for each class (frames from segments that were human-labelled as having or not having the class). As this is a relatively small dataset, several thousand additional negative frames were randomly sampled from the large training set from videos which are not labelled with the class in question.

It is important to note that while we could have sampled additional negatives (each class has approx. 1 billion ``probably negative" frames in the entire dataset), this was not done as these frames tend to be trivial for a model to classify, distracting from the real problem of distinguishing between positive and negative segments \textit{within the same video} - which was found experimentally to be much more difficult for a model to do.

Each frame descriptor from this dataset $X$ is then passed through the trained video-level model (as if it was a video, to get a 1000-class prediction vector) and the results are concatenated to form a new feature vector $\{X, fl(X)\} \in \mathbb{R}^{2252}$. This forms our frame-level dataset.

\subsection{Gradient Boosted Trees Model}

After this frame-level dataset has been constructed (around $5000\times2252$), for each class $c$ a set of XGBoost\cite{xgboost} models $xgb_c$ is trained with a 5-fold cross-validation approach, for a total of 5000 models (each model takes a few seconds to train - about 24 hours total). Because each class only has on average $125$ positively labelled training segments from which the entire training set is derived, there is a large variance between folds which use different segments for validation. Hence, at test time, the average of these 5 models for each class are taken to improve performance.

\subsection{Model composition}
\label{sec:modelcomposition}

The XGBoost model $xgb_c$ has been trained on a specially crafted (and thus \textbf{biased}) subset of the full set of frames. This is because the Youtube-8M segments data only has segments labelled for videos \textbf{where the video itself has the class $c$} - the training data is very heavily biased towards these videos.

One solution to this problem would be to construct a frame dataset which accurately reflects the distribution of frames or l in the whole dataset, this however presents two challenges:

\begin{enumerate}
    \item Very few frames in the dataset are labelled (on a segment level), so creating a representative data would be difficult - it is not as simple as just subsampling the data.
    \item This has the risk of transforming the problem into one of mostly detecting which videos have the class versus which videos are completely unrelated, instead of the more difficult  (and important) task of distinguishing positive and negative frames \textit{within videos that are labelled with the class}.
\end{enumerate}

\noindent Instead, we model the output of $xgb_c$ as a conditional probability:

\begin{equation}
\label{eq:prob1}
xgb_c(\{X, fl(X)\}) = P(X \in c \mid V_X \in c)
\end{equation}

\noindent where $V_X$ is the video of frame $X$. Essentially, our XGBoost model predicts the probability that a frame $X$ has class $c$, given that the video the frame was from has class $c$.

Our goal is to calculate the probability $P(X \in c)$. We can use Bayes' rule on (Eq. \ref{eq:prob1}) to get

\begin{equation}
\label{eq:prob2}
P(X \in c \mid V_X \in c) = \frac{P(V_X \in c \mid X \in c) \cdot P(X \in c)}{P(V_X \in c)}
\end{equation}

\noindent $P(V_X \in c \mid X \in c)$ is 1, as a video of segment in class $c$ must necessarily also be in class $c$. And so:

\setlength{\jot}{2ex}
\begin{equation}\begin{gathered}
    \label{eq:prob3}
    P(X \in c \mid V_X \in c) = \frac{P(X \in c)}{P(V_X \in c)} \\
    P(X \in c) = P(X \in c \mid V_X \in c) \cdot P(V_X \in c) \\
    P(X \in c) = xgb_c(\{X, fl(X)\}) \cdot P(V_X \in c)
\end{gathered}\end{equation}

\noindent Now we just need the video-level probability $P(V_X \in c)$. Luckily, we can use the state-of-the art video-level model\cite{2018best} from Section \ref{sec:baseline} to estimate a video-level probability $vl_c(V_X) = \cdot P(V_X \in c)$. The final prediction for each frame is the product of this video-level model and the XGBoost model - and the prediction for each 5 second segment is this probability averaged over the 5 frames in the segment.

\begin{equation}
    P(S \in c) = \frac{1}{5} \sum_{i=1}^{5} xgb_c(\{X_i, fl(X_i)\}) \cdot vl_c(V_S)
\end{equation}

\noindent where $S$ is a segment containing frames $X_1 ... X_5$. By averaging the likelihoods of the 5 frames in each segment, we can consider more information for our prediction instead of just using information from a single frame.

\subsection{Model weighting}

The product of the two models, in a sense, `equally weights' the class probability predictions estimated by our XGBoost model and the video-level model. Where both models are perfect predictors on their respective datasets, this works fine (as the equations show) - however, the reality is that some models are more effective than others, and so it may be useful to change these weightings.

The product can also be interpreted as a geometric mean ensemble between the two models (as the geometric mean is just the product followed by the square root, which would not affect the ranking of segments for MAP). Thus, we can introduce an additional parameter $p$ to control the weighting between the models

\begin{equation}
    P(X \in c) = \left(xgb_c(...)\right)^p \cdot vl_c(V_S)
\end{equation}

\noindent A $p > 1$ provides higher weight to XGBoost, while $p < 1$ provides higher weight to the video-level model. A higher $p$ in general produces smaller (skewed) probabilities to be predicted, but as MAP only cares about \textit{ranking} of segments, there is no need to correct for this skew. This can be seen as analagous to weighting models in a linear ensemble: $p \cdot \left(xgb_c(...)\right) + \cdot vl_c(V_S)$.

Experimentally, $p=4$ was found to be a good value, improving the MAP of this model from 0.77552 to \textbf{0.79545}.



\section{Deep learning for event localization}
\label{sec:deeplearning}

\subsection{5-frame models}

As the goal is to label $5$ second segments, we built models that process 5 frames at a time (a single segment) without considering other information from elsewhere in the video. 

In particular, 2 models were trained based on the previously successful NetVLAD\cite{arandjelovic2016netvlad,2017best} network architecture and one based on a deep bag of frames (DBoF)\cite{2016dataset} network architecture. The models were first pre-trained on the 2nd Youtube-8M dataset, by selecting 5 frames from the videos at random, and assuming that these segments inherit the classes of the video - yielding a large but very approximate dataset. Finetuning on specifically annotated 5 second sequences followed. In the finetuning process any unlabelled classes (neither positively or negatively, due to the extremely sparse labelling) were masked out and not trained on, to avoid affecting the model learning process with incorrect labels. Furthermore, initial pretraining on the bigger dataset was carried out for longer ($\approx50K$ and $\approx10K$ steps at a batch size of $1024$ and $2048$ for the VLAD and DBoF models respectively) than finetuning ($\approx500$ iterations at previously mentioned batch sizes). This was done to avoid overfitting to the comparatively tiny segment-annotated dataset.

Single VLAD model achieved MAP of $\approx0.74$ and combining the three models improved performance further. Empirically we observed that multiplying probabilities (geometric mean) gave a better MAP uplift than averaging (arithmetic mean) the model predictions (Table~\ref{table1:deepperformance}).

\subsection{Global model}
A 5 second snippet does not give all the possible useful information about a video - context is also important; for example, if  a video is all about cars, then it is more likely that a certain given segment is also about cars. For this reason, we want to leverage `global` information from the video, where just relying on $5$ frames and ignoring broader video context might result in misclassification and lead to suboptimal performance. 

To tackle this, we combined the segment-level model outputs with label predictions for the whole video. This was done in a similar way to the Section \ref{sec:modelcomposition}; a mean was taken between the local predictions of the 5-frame models and the . It was found that the geometric mean significantly outperformed the arithmetic mean for combining these models (Table~\ref{table1:deepperformance}).

\subsection{Localization model}

\begin{figure}[h]
 \centering
 \includegraphics[width=0.48\textwidth]{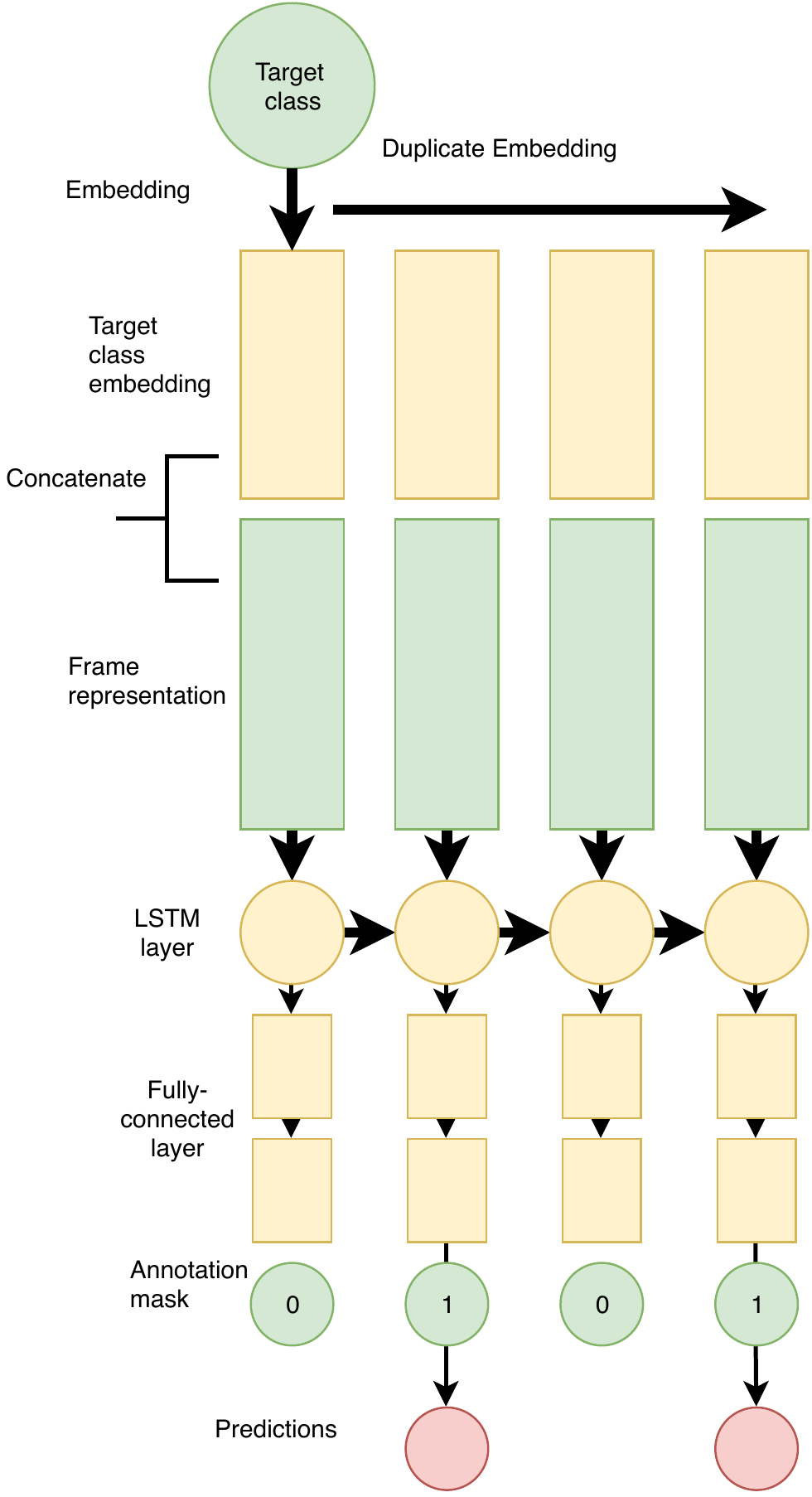}
    \caption{Schematic representation of the localization network. Inputs are colored green, while latent values and outputs are colored yellow and red, respectively.}
    \label{fig:locnetwork}
\end{figure}

The model described in previous section helped with overall predictive performance. However, the context modelling video-level model did not add any temporal differentiation. Hence, we introduced a localization model to help prioritize sections of video that contain the target classes.

The localization model was implemented as follows. The target class is passed through an embedding layer converting our sparse one hot encoding of $1000$ classes to a low 32-dimensional continuous representation. This representation is then concatenated with the 1152-dimensional frame-level representation, before being passed to an dynamic length bidirectional long short-term memory (LSTM)\cite{hochreiter1997long} network. Next, the LSTM output is processed by a fully-connected layer to yield a single logit per frame. Finally, the sigmoid activation is applied to yield class probabilites and this is trained with backpropagation. Importantly, unannotated segments were masked out of the loss calculation not to interfere with model training via backpropagation. Schematic representation of this network is shown in Figure~\ref{fig:locnetwork}. During inference all videos were passed through the network in combination with all $1000$ target classes in order to get class specific probabilities.

\subsection{Combining the models}

\begin{table}[h]
\begin{tabular}{lr}
\toprule
Models                                       & MAP score \\ 
\midrule 
VLAD Model (5-frame)                         & 0.74056   \\
3x 5-frame (arithmetic mean)                 & 0.76124   \\
3x 5-frame (geometric mean)                  & 0.76523   \\
3x 5-frame, global model                     & 0.78209   \\
3x 5-frame, global model, localization model & \textbf{0.78732}  \\
\bottomrule
\end{tabular}
\vspace{0.1cm}
\caption{MAP score for different model or model combination predictions. MAP scores are reported based on the private Kaggle leaderboard.}
\label{table1:deepperformance}
\end{table}

Final score ($S_{v,s,c}$) for a segment $s$ of video $v$ for class event $c$  was calculated as follows:
\begin{equation}
S_{v,s,c} = p_{video} \cdot p_{loc} \cdot (p_{VLAD1} \cdot p_{VLAD2} \cdot p_{DBoF}) ^{2/3}.
\end{equation}

In the equation $p_{video}$ are predictions for video, $p_{loc}$ are localization model predictions and 5-frame models predictions are represented as  $p_{DBoF}$ and $p_{VLAD}$. Video model gives class predictions for each video, while $5$-frame models and localization model give class predictions for each segment in video. For each target class the segments were sorted by final score in descending order and top ranking segments were reported.

We would like to note that due to time and computational constrains the models were not (exhaustively) tuned. Rather, they serve as a proof of concept and with proper tuning the performance can be further improved.

\section{Final meta-model}

\begin{table}[h]
\begin{tabular}{lr}
\toprule
Model                                       & MAP score \\ 
\midrule 
Video-Level Baseline (Sec. \ref{sec:baseline})   & 0.68182   \\
Gradient Boosting Approach (Sec. \ref{sec:gbm})  & 0.79545   \\
Deep-learning approach (Sec. \ref{sec:deeplearning})   & 0.78732   \\
Rank Average                     & \textbf{0.80459}   \\
\bottomrule
\end{tabular}
\vspace{0.1cm}
\caption{MAP score for different model or model combination predictions. MAP is reported based on the Kaggle private leaderboard score.}
\label{table2:final}
\end{table}

Predictions from the models described in previous two sections were, ensembled into final predictions. This was done using a rank average: for each class, the predictions from the two models are assigned a score based on their rank in the prediction list, this rank is averaged, and then the list is sorted on the average rank in all the models. This allows the models to be ensembled without havig to keep around the full table of probabilities, and is robust to when the probability distributions of the two models are very different (as is likely the case here). The ensembled prediction achieved MAP score of \textbf{0.80459} (Table~\ref{table2:final}) and ranked 5th as part of the 3rd Youtube-8M video recognition challenge. 
Code to reproduce the results presented here, including the chosen hyperparameters, is available at \url{https://github.com/mxbi/youtube8m-2019}.

{\small
\bibliographystyle{ieee_fullname}
\bibliography{egbib}
}

\end{document}